\documentclass[conference]{IEEEtran}
\IEEEoverridecommandlockouts

\usepackage{hyperref}
\usepackage{amsmath,amssymb,amsfonts}
\usepackage{graphicx}
\usepackage{textcomp}
\usepackage{algorithm}
\usepackage{algpseudocode}
\usepackage{multirow}
\usepackage{comment}
\usepackage{graphicx}
\usepackage{amsmath}
\usepackage{cite}
\usepackage{comment}
\usepackage{subcaption}
\usepackage{algpseudocode}
\usepackage{amsmath}
\usepackage{comment}
\usepackage{multirow}
\usepackage{booktabs}
\usepackage{balance} 
\usepackage{array}
\usepackage[utf8]{inputenc}
\usepackage{xcolor}
\usepackage{url}
\usepackage{xurl}
\usepackage{authblk}  
\usepackage{float} 
\usepackage{cite}
\usepackage{enumitem}
\usepackage[normalem]{ulem}
\def\BibTeX{{\rm B\kern-.05em{\sc i\kern-.025em b}\kern-.08em
    T\kern-.1667em\lower.7ex\hbox{E}\kern-.125emX}}
\begin{document}

\title{Adaptive Composition of Machine Learning as a Service (MLaaS) for IoT Environments}

\author[1]{Deepak Kanneganti}
\author[1]{Sajib Mistry}
\author[1]{Sheik Mohammad Mostakim Fattah}
\author[1]{Aneesh Krishna}
\author[2]{Monowar Bhuyan}
\affil[1]{\textit{School of EECMS, Curtin University, Australia}}
\affil[2]{\textit{Department of Computing Science, Umeå University, Sweden}}
\affil[ ]{\{s.kanneganti, sajib.mistry, sheik.fattah, a.krishna\}@curtin.edu.au, monowar@cs.umu.se}

\maketitle
\author{
Deepak Kanneganti, Sajib Mistry, Sheik Mohammad Mostakim Fattah, Aneesh Krishna, and Monowar Bhuyan
\thanks{Deepak Kanneganti, Sajib Mistry, Sheik Mohammad Mostakim Fattah, and Aneesh Krishna are with the School of Electrical Engineering, Computing and Mathematical Sciences, Curtin University, Australia (emails: \{s.kanneganti, sajib.mistry, sheik.fattah, a.krishna\}@curtin.edu.au).}
\thanks{Monowar Bhuyan is with the Department of Computing Science, Umeå University, Sweden (email: monowar@cs.umu.se).}
}

\begin{abstract}

The dynamic nature of Internet of Things (IoT) environments challenges the long-term effectiveness of Machine Learning as a Service (MLaaS) compositions. The uncertainty and variability of IoT environments lead to fluctuations in data distribution, e.g., concept drift and data heterogeneity, and evolving system requirements, e.g., scalability demands and resource limitations. This paper proposes an adaptive MLaaS composition framework to ensure a seamless, efficient, and scalable MLaaS composition. The framework integrates a service assessment model to identify underperforming MLaaS services and a candidate selection model to filter optimal replacements. An adaptive composition mechanism is developed that incrementally updates MLaaS compositions using a contextual multi-armed bandit optimization strategy. By continuously adapting to evolving IoT constraints, the approach maintains Quality of Service (QoS) while reducing the computational cost associated with recomposition from scratch. Experimental results on a real-world dataset demonstrate the efficiency of our proposed approach.
\end{abstract}

\begin{IEEEkeywords}
MLaaS, Internet of Things, Service Composition, Combinatorial Optimization, Contextual Multi-Armed Bandit
\end{IEEEkeywords}

\section{Introduction}

\IEEEPARstart{T}{he} rapid evolution of the IoT has transformed industries by enabling edge devices such as sensors and actuators to collect and exchange real-time data. Sectors including healthcare, manufacturing, and smart cities leverage IoT for real-time data acquisition and remote monitoring \cite{lee2015internet}. Recently, machine learning (ML) has enabled IoT systems to recognize patterns in data and predict future outcomes. The integration of ML has further advanced IoT applications to provide intelligent decision-making. For instance, in healthcare, ML and IoT facilitate human activity recognition (HAR) by detecting and monitoring patient movements \cite{bianchi2019iot}.

Machine Learning as a Service (MLaaS) is a cloud-based service that offers an ML environment and infrastructure to enable users to build and deploy ML services \cite{ribeiro2015mlaas}. The increasing \textit{data volume} and \textit{cost of ML operations} have made MLaaS an essential component of IoT environments. Tech giants such as Amazon, Google and Microsoft offer MLaaS solutions, including \textit{AWS}\footnote{https://aws.amazon.com/machine-learning/}, \textit{Google Cloud AI}\footnote{https://azure.microsoft.com/en-us/services/machine-learning/}, and \textit{Azure}\footnote{https://aws.amazon.com/sagemaker/pricing/} through cloud marketplaces \cite{noshiri2021machine}. These services are distinguished by functional attributes (e.g., model and training data specifications) and non-functional expressed through QoS attributes (e.g., accuracy, reliability, and latency).

The composition of MLaaS enables IoT providers to integrate multiple MLaaS services to form a unified service that meets specific functional and non-functional requirements \cite{xie2024skyml}. For example, HAR applications may require services that accommodate diverse user groups, sensor modalities, environmental conditions, and activity types. Given that IoT applications often require \textit{ high accuracy} and \textit{low latency}~\cite{fadlullah2018delay}, a single MLaaS service may not fully meet these requirements. In such cases, IoT providers may \textit{compose multiple MLaaS services} to leverage their unique strengths, ensuring a more robust and adaptive HAR framework. For example, a smart fitness application may integrate multiple MLaaS solutions where one service focuses on sensor-based movement recognition, another on vision-based activity detection, and a third on low-power edge classification. By composing these services, the system can maintain high accuracy across different environments while ensuring low latency for real-time feedback.

IoT environments are \textit{inherently dynamic} due to continuous changes in user behavior, sensor data distributions, and evolving system requirements~\cite{fadlullah2018delay}. This dynamism introduces challenges in maintaining the effectiveness of an MLaaS composition over time. The composition may become \textit{ineffective} due to the \textit{uncertainty} and \textit{variability} in IoT environments~\cite{muccini2020leveraging}. For example, an IoT provider may initially compose MLaaS services from Amazon SageMaker, Google Cloud AI, and Microsoft Azure to perform HAR with a personal QoS preference of \textit{high accuracy} and \textit{low latency}. Although the composition may perform optimally at first, its effectiveness often degrades over time as it encounters shifts in user activity patterns or latency fluctuations. In such cases, the provider must either \textit{recompose} or \textit{update} the existing composition. Recomposing involves repeating the entire composition process, which makes it \textit{computationally expensive} and \textit{time-consuming}.

To address these challenge, we focus on \textit{adaptive MLaaS composition}, which dynamically adjusts an existing MLaaS composition in response to changing IoT conditions. Instead of recomposing from scratch, adaptive MLaaS composition applies real-time monitoring and incremental updates to maintain optimal performance. By continuously refining the composition based on evolving data and system constraints, the adaptive MLaaS composition ensures that IoT applications consistently meet their QoS requirements, making them more resilient in dynamic IoT environments.

Traditional approaches to adaptive service composition, such as goal-driven strategies, rule-based methods, and QoS-based adaptation, have been widely employed to achieve dynamic service composition \cite{ma2011adaptive, wang2010adaptive, ardagna2007adaptive}. These methods have proven effective in domains where service requirements and environmental constraints are relatively stable. However, they are inadequate for the adaptive MLaaS composition in IoT environments due to the dynamic and heterogeneous nature of IoT-generated data. Unlike conventional services, MLaaS composition involves real-time variability across data streams, learning models, and system constraints, a challenge that static rules and QoS attributes alone cannot effectively address. For instance, selecting an MLaaS service solely based on QoS attributes such as accuracy and latency may overlook fundamental incompatibilities in training data specifications, model architecture, and computational constraints, leading to suboptimal learning outcomes. The heterogeneity of IoT data, influenced by varying sensor accuracies, network conditions, and data distribution shifts, further complicates the problem. Conventional rule-based or QoS-driven methods lack the necessary flexibility to adjust to these changes dynamically, limiting their effectiveness in ensuring optimal model performance and reliability \cite{ma2011adaptive, ardagna2007adaptive}. We identify the four key challenges to perform adaptive MLaaS composition for an IoT environment.

\vspace{-1mm}
\begin{itemize}[itemsep=0ex, leftmargin=2ex]
    \item \textit{Identification:}  The primary challenge is identifying underperforming services within a given composition. For example, let us consider three MLaaS services that are composed to perform a HAR. Some of the services perform poorly over time due to the uncertainty and variability of the IoT environment. Traditional approaches rely solely on QoS parameters, such as accuracy, but often fail to capture underlying factors that affect service performance \cite{song2019profit, wang2019measure}. This highlights the need to consider both functional and QoS attributes for a more reliable assessment.
    \item \textit{Candidate MLaaS Selection:} The rapid growth of MLaaS services in the cloud marketplace makes the selection of MLaaS complex. Let us consider that the number of potential MLaaS services can be significantly large. Considering each MLaaS service to update the existing composition may not always be practical. Instead, an efficient selection strategy is required to identify the potential candidate services.
    \item \textit{Adaptive Composition:} Given a set of potential candidate services, the challenge is to determine the most suitable service to update the composition. For instance, Let us consider an MLaaS composition that consists of three services, \( M_C = \{M_1, M_2, M_3\} \). Consider \( M_3 \) is underperforming and needs to be updated with a candidate service \( M_x \). However, determining whether \( M_x \) is the most suitable service to update the composition is not straightforward \cite{zhao2018federated}. \( M_x \) may be composable if its attributes, including model heterogeneity, architecture, and training specifications, align with the existing composition. Despite being functionally composable, the QoS attributes such as \textit{latency, accuracy, and resource efficiency} can only be validated after recomposition composition\cite{lai2021oort}. However, recomposition of the entire service is often both \textit{computationally expensive} and \textit{time-consuming}. This highlights the need for an efficient mechanism to \textit{determine adaptive composition}.
    \item \textit{Optimization:} Consider a scenario where multiple MLaaS services are underperforming, each with multiple replacement candidates. In this case, the number of potential compositions grows exponentially, making the selection process highly complex. This highlights the need for an optimization mechanism that strategically streamlines the process and selects the most effective replacements.


\end{itemize}

To address this challenge, we propose a new \textit{adaptive MLaaS composition framework} that dynamically updates service compositions to enhance efficiency and reliability. We build a \textit{service assessment model} that evaluates ML services based on their functional and QoS attributes. This assessment enables the identification of underperforming services that may degrade the overall composition. Next, we introduce a \textit{candidate service selection model} that filters and ranks potential replacements using real-time monitoring and historical performance data. This model ensures that only the most reliable and contextually suitable services are selected, maintaining an optimal MLaaS composition. To further improve adaptability, we propose an \textit{adaptive MLaaS composition algorithm} that evaluates candidate services in real time, ensuring seamless integration with minimal disruption. Furthermore, we introduce a \textit{multi-armed bandit (MAB) algorithm} that optimizes service selection by intelligently exploring and exploiting available service options using feedback-driven learning. This approach reduces search complexity and enhances the robustness of MLaaS compositions.


\section{Related Work}
\subsection{MLaaS Selection and Composition in IoT}
MLaaS has gained significant attention in recent years. Ribeiro et al. \cite{ribeiro2015mlaas} introduced MLaaS as a cloud-based service providing pre-trained models and computational resources, enabling seamless ML integration. MLaaS is used across various IoT environments to reduce cost and enable intelligent decision-making. For instance, \cite{tay2019xylorix} leveraged MLaaS for urban planning through Urban Modeling as a Service (UMaaS), while in healthcare, \cite{pohl2018data} proposed a Risk Prediction as a Service, aiding professionals in assessing patient health risks. However, relying on a single MLaaS provider can lead to dependency issues and limit performance benefits. To address this, Xie et al. \cite{xie2024skyml} propose an MLaaS federation that combines multiple MLaaS, allowing systems to avoid vendor lock-in and leverage the strengths of different services. 

ML selection is critical for evaluating and optimizing performance in collaborative learning (CL), where data and model heterogeneity can influence the effectiveness of ML models over time. Traditional approaches, such as the Shapley value method \cite{song2019profit}, evaluate client contributions but suffer from high computational costs, limiting scalability. Similarly, the Leave-One-Out (LOO) approach \cite{wang2019measure} evaluates contributions based on accuracy but overlooks other QoS factors. In FL, the Normalized Contribution Score (NCS) \cite{zhang2024towards} considers weight updates but disregards data quality and system constraints. While existing CL methods assess client contributions through functional or non-functional attributes, they overlook their combined impact. We focus on evaluating service performance within an MLaaS composition, integrating both aspects for a balanced evaluation of service contributions.

The problem of client selection has been widely studied, with research focusing on data, model, and system heterogeneity. Early approaches emphasized evaluating data distributions to guide model selection, ensuring that services with similar characteristics are effectively integrated \cite{zhao2018federated, huang2022tackling}. Other techniques assessed training dynamics, where client contributions were measured through model weights and gradient updates \cite{fu2023client}. Additionally, prioritizing models based on data representativeness and system efficiency has improved time-to-accuracy in heterogeneous environments \cite{lai2021oort}. More recent work has extended adaptive FL section frameworks by incorporating dynamic factors such as reliability, resource availability, and cost, optimizing service integration \cite{zhang2023multi}. By building on these principles, our approach synchronizes functional and non-functional attributes of MLaaS services to determine adaptive composition.

\subsection{Adaptive Service Composition and Optimization}
Service composition facilitates seamless integration across cloud computing, MLaaS, IoT, web, and social networks. Given the dynamic nature of these environments, adaptive composition has emerged as a key research area. Early work proposed monitoring-based adaptation frameworks \cite{ma2011adaptive} and reinforcement learning (RL)-driven approaches, such as Q-learning, to enable dynamic composition \cite{wang2010adaptive, wang2014adaptive}. Ardagna et al. \cite{ardagna2007adaptive} introduced a multi-channel adaptive information model for flexible service selection, while Tan et al. \cite{tan2014automated} employed genetic algorithms to optimize QoS through recovery-based adaptation. Additionally, Wang et al. \cite{wang2014adaptive} leveraged multi-agent RL and game theory to optimize adaptive composition. The method focuses on static rules, predefined strategies, and Q-learning to identify candidate services for adaptive composition. However, these approaches fail to address real-time heterogeneity in data, models, and system constraints. Our work addresses these challenges by enabling adaptive MLaaS service composition that dynamically incorporates evolving data and system conditions.

\begin{figure*}[]
    \centering
    \includegraphics[width=0.9\textwidth]{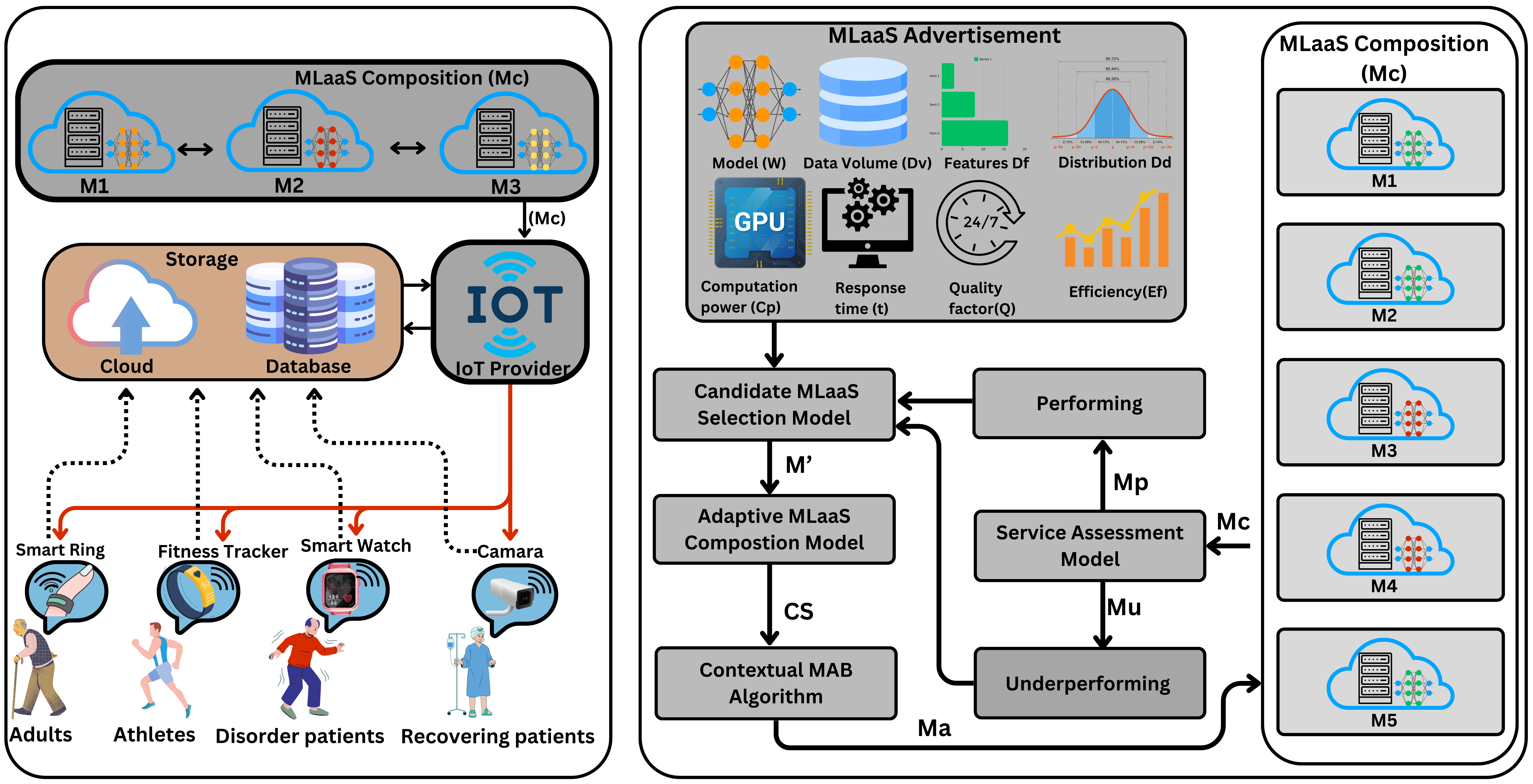}
    \caption{Illustration of the Adaptive MLaaS Composition: (a) Motivation Scenario and (b) Proposed Framework}
    \label{fig1}
    \vspace{-5mm}
\end{figure*} 

\section{Motivation Scenario}
Let us consider an IoT provider that aims to develop a smart healthcare system that requires Human Activity Recognition (HAR). The provider prefers an MLaaS service that can accommodate clients with diverse age groups, sensor modalities, medical conditions, and recovery stages. Let us assume that the provider has performed MLaaS composition ($M_c$) to create a unified HAR service to accommodate diverse clients, as shown in Fig. \ref{fig1}a. Due to uncertainty and variability in IoT environments, IoT environments are inherently dynamic. The performance of the MLaaS composition may deteriorate over a period of time. Hence, the IoT provider requires an effective adaptive composition strategy to update the existing composition to meet the desired requirement.



The primary challenge is to identify MLaaS services that perform poorly within the given composition ($M_c$), where one or more services may perform poorly. We assume that both functional and non-functional attributes influence the underperforming services. Hence, the IoT provider requires an effective evaluation method to determine underperforming components. Once identified, the next challenge is to select potential candidate services from the MLaaS marketplace. The selection process is non-trivial, as thousands of services advertise their attributes. Hence, the IoT provider needs an effective candidate selection strategy to filter potential candidates.

Let us assume that a candidate MLaaS service is being filtered to update the composition. Evaluating the adaptive composition of a candidate service with the existing composition is not straightforward. Functional attributes, such as model parameters and training data specifications, must align with the existing composition. Evaluating the QoS of the updated composition is only possible after recomposition, which is computationally expensive and time consuming. Therefore, the provider needs an adaptive and reliable evaluation mechanism to determine the adaptive composition of the candidate service to update the composition.

Consider that multiple MLaaS services are underperforming, each with 10 potential MLaaS services to update the composition \( M_c \). Each underperforming service can be replaced by any of these options, leading to \( 10^n \) possible combinations, where \( n \) is the number of services being replaced. For \( n = 2 \), this results in 100 combinations. As \( n \) increases, the number of possibilities grows exponentially, making exhaustive evaluation impractical. Hence, an efficient optimization technique is needed to reduce computational overhead and ensure scalability. To address these challenges, we propose an adaptive MLaaS composition framework to identify underperforming services and update the composition.

\section{Key Definitions and Problem Statement}\label{MLaaS_problem}
This section introduces the key concepts and definitions for understanding the adaptive MLaaS composition problem.

\noindent\textbf{Definition 1 (Formal MLaaS).} 
\textit{A Formal MLaaS service $M$ is represented by the tuple $M_ = \langle id, F, E_f, Q, t, R \rangle$, where:}
\begin{itemize}[itemsep=0ex, leftmargin=2ex]
    \item $id$ is the unique MLaaS service ID.
    \item $F$ is the functional specification offered by $M$, which includes the model specifications ($\mathcal{N}_{\Phi}$) (e.g., weights $w$ and gradients $\nabla w$) and data specifications ($D_s$) (e.g., volume $D_v$, modality $D_m$, and features $D_f$).
    \item $E_f$ represents evaluation metrics, assessing the performance of $M$ using indicators such as accuracy, $R^2$, and task-specific evaluation scores.
    \item $Q$ is the quality factor, reflecting the historical service performance and reliability over multiple executions.
    \item $t$ is the response time, indicating the latency of $M$ in processing client requests.
    \item $R$ is the reliability score, evaluating the consistency of $M$ in delivering stable results.
\end{itemize}

\noindent\textbf{Definition 2 (Service Contribution Score (SCS)).} 
The SCS is used to identify the underperforming service from a given composition. The SCS of an  \( M_c \) is determined using the following equation:
\begin{equation}\label{SCS}
SCS^{(M_c)} = \alpha \cdot QoS_{M_c}  + \beta \cdot f^{(c)}
\end{equation}
where:  
\begin{itemize}[itemsep=0ex, leftmargin=2ex]
    \item \( \alpha \) and \( \beta \) are weighting parameters that balance the influence of two key components.  
    \item \( QoS \) represents the \textit{QoS-based contribution score}, which quantifies the marginal impact of each MLaaS service. 
    \item \( f^{(c)} \) is the \textit{Functional contribution score}, which measures how closely each MLaaS is associated with the composition.
\end{itemize}

\noindent\textbf{Definition 3 (Adaptive MLaaS Composition Problem).}  
The adaptive MLaaS composition problem is to determine the adaptive composition between a candidate service \( M_i \) and the existing composition \( M_c \). To address this, the adaptive composition model is designed to generate a confidence score (CS) and is defined as :
\begin{equation}\label{conf_score}
    \text{CS}(M_p^c, M_i) =
    \begin{cases}
        1, & \text{if } M_i \text{ aligns with } M_c, \\
        0, & \text{otherwise}.
    \end{cases}
\end{equation}
The confidence score [0,1] determines the composition of a candidate MLaaS service with \( M_c \), where 1 indicates full alignment and 0 signifies incompatibility.

\section{Adaptive MLaaS Composition framework}\label{section 3}

The proposed adaptive composition framework consists of five key components, as illustrated in Fig. \ref{fig1}. The first component is the cloud marketplace, where MLaaS providers advertise their functional and non-functional attributes. The second is the existing MLaaS composition, which includes the initial set of services considered by the IoT provider. The third component, a service assessment model, helps identify underperforming services in the composition. Later, a candidate service selection model is developed to filter potential services. The fifth component is the adaptive MLaaS composition model, which determines whether candidate services can integrate with the existing composition. Finally, a contextual MAB algorithm is employed to perform optimization and avoid exhaustive exploration of all possible compositions.


\subsection{Service Assessment Model}

The service assessment model evaluates the contributions of individual services within an existing composition. Due to the dynamic nature of IoT, these contributions may change over time. Traditional methods like Leave-One-Out (LOO) \cite{wang2019measure} and Shapley value-based assessment (SHAP) \cite{song2019profit} focus mainly on accuracy while ignoring key QoS parameters such as latency, quality, and reliability. Our approach addresses this limitation by incorporating multiple QoS aspects for a more comprehensive evaluation. The contribution of each service is determined using the QoS score formula:


\begin{equation}
    Q(M) = \frac{1}{n} \sum_{p \in \{E_f, Q, t, R\}} Q_p(M)
\end{equation}

Where, \( E_f \)  represents the effectiveness factor,  \( Q \)   denotes the quality factor,  \(t\)  refers to computation time (latency), and \( R \)  indicates the reliability score. These factors collectively define the system’s overall quality. To measure the individual contribution of a service \( i \) within the composition   \( M_c \), we quantify its impact on the overall QoS by measuring the change in the QoS score upon its removal. This impact is then normalized to ensure fairness and proportionality among all services. The QoS contribution score is defined as:

\begin{equation}\label{qos_3}
QoS_{M_c} = \frac{Q(M_c) - Q(M_c \setminus i)}
{\sum_{j \in M_c} \left( Q(M_c) - Q(M_c \setminus j) \right)} \times 100
\end{equation}

Incorporating ML specifications (e.g., weights and gradients) can effectively capture the true impact of services in a composition. Traditional methods, like SHAP and Leave-One-Out (LOO), primarily focus on accuracy-based evaluations. These techniques may often fail to fully capture the true impact of a service within a composition. To address this limitation, we leverage techniques from the FedTruth \cite{ebron2023fedtruth} framework to assess service contributions in a more refined and systematic manner. ML specifications can determine a service’s influence on the learning process and a better understanding of each role of the service. This method goes beyond binary contribution assessments by incorporating model weights, thereby capturing both the presence and the effectiveness of each service’s contribution to overall performance. The functional contribution score for each service \( i \) is given by:




\begin{equation}
\small S_t^{(c)} = \frac{d(W_t^i - W_t, W_t^{(c)} - W_t)}
{\sum_{j=1}^{n_t} d(W_t^i - W_t, W_t^{(j)} - W_t)} ,\quad
g\left(S_t^{(c)}\right) = \frac{1}{S_t^{(c)}}
\end{equation}


The function \( d(\cdot) \) quantifies the distance between model updates using Euclidean or angular distance. Specifically, \( W_t^i \) represents the distance between an individual service’s weight and the update, while \( W_t^C \) measures the distance between the composed service and the update. A smaller \( S_t^{(c)} \) indicates a stronger alignment between the service and the composition, meaning it has a greater impact. The function \( g(S_t^{(c)}) \) assigns a weight based on this deviation, ensuring that services with closer alignment contribute more significantly. The functional contribution score is then defined as:


\begin{equation}\label{net contribution score}
f^{(c)} = \frac{g(S_t^{(c)}) \cdot S_t^{(c)}}
{\sum_{j \in S_t} g(S_t^{(j)}) \cdot S_t^{(j)}}
\vspace{-2mm}
\end{equation}

\begin{algorithm}
\caption{Service Assessment Model}\label{alg1}
\begin{algorithmic}[1]
\State \textbf{Input:} \(M_c=\{w,E_f, Q, t, R_t \}, M,W_t, \alpha, \beta, \theta \) 
\State \textbf{Output:} \( SCS, M_u \)
\State Compute overall QoS score:
\State $Q(M) \gets \frac{1}{n} \sum_{p \in \{E_f, Q, t, R\}} Q_p(M)$  
\For{$i \in M_c$}  
    \State $QoS_{M_c}(i) \gets \frac{Q(M_c) - Q(M_c \setminus \{i\})}{\sum_{j \in M_c} (Q(M_c) - Q(M_c \setminus \{j\}))} \times 100$  
\EndFor
\For{$i \in M_c$}  
    \State $S_t^{(i)} \gets \frac{d(W_t^i - W_t, W_t^{(c)} - W_t)}{\sum_{j \in S_t} d(W_t^j - W_t, W_t^{(j)} - W_t)}$
    \State $g(S_t^{(i)}) \gets \frac{1}{S_t^{(i)}}$
    \State $f^{(i)} \gets \frac{g(S_t^{(i)}) \cdot S_t^{(i)}}{\sum_{j \in S_t} g(S_t^{(j)}) \cdot S_t^{(j)}}$
\EndFor
\For{$i \in M_c$}  
    \State $SCS_i \gets \alpha \cdot QoS_{M_c}(i) + \beta \cdot f^{(i)}$  
    \If{$SCS_i < \theta$}  
        \State Append $i$ to $\mathbf{M_u}$  
    \EndIf
\EndFor
\State \Return $\mathbf{SCS}, \mathbf{M_u}$
\end{algorithmic}
\end{algorithm}

Equation \ref{SCS} formulates the Service Contribution Score (SCS) by combining both the QoS and functional contribution scores. \textit{Definition 2} provides an overview of this formulation, which is key to determining the contributions of individual services. A higher SCS indicates a more significant contribution within the composition. Algorithm \ref{alg1} demonstrates a service assessment model designed to identify underperforming services. The algorithm begins with the input of an MLaaS composition \( M_c \) and parameters \( \alpha \), \( \beta \), and \( \theta \), which are used to determine service contributions (\textit{Algorithm \ref{alg1}, lines 1–3}). The evaluation consists of two stages for a comprehensive assessment. In the first stage, the QoS-based contribution score is computed by removing each service sequentially and measuring its impact using Equation~\ref{qos_3} (\textit{Algorithm \ref{alg1}, lines 4–7}). The second stage measures the functional contribution by assessing the distance between each service and the composition using Equation~\ref{net contribution score} (\textit{Algorithm \ref{alg1}, lines 8–12}). The final Service Contribution Score (SCS) is computed using Equation~\ref{SCS} to balance functional and QoS attributes. Services that closely align receive higher scores, while those with low score are classified as underperforming. The threshold \( \theta \) (\( \theta > 10\% \)) is used to filter underperforming services (\( M_u \)), identifying those that fail to meet the criterion as outputs (\textit{Algorithm \ref{alg1}, lines 13–19}).

\subsection{Candidate MLaaS Selection Model}
The candidate MLaaS selection model identifies the potential candidates to update the MLaaS Composition. Algorithm~\ref{alg2} provides the detailed workflow for selecting potential MLaaS services. The algorithm takes a set of MLaaS services \( M \) as input, along with a subset of underperforming services  \( M_u \) identified by the service assessment model \ref{alg1}.The model filters candidate services and provides them as output \( M' \). In the first phase, the model performs composition among underperforming services \( M_u^c \) (\textit{Algorithm \ref{alg2}, lines 1–4}). \( M_u^c \)  is used to determine the functional attributes needed to select candidates that can integrate seamlessly into the existing composition\( M_c \). Functional attributes are used to filter out potential candidate MLaaS services by assessing their functional relevance to underperforming services $M_u$. Services that satisfy these conditions are filtered using the following conditions (\textit{see Algorithm \ref{alg2}, lines 5–8}). \( \mathcal{N}_{\Phi} \) represents the model specifications, \( D_m \) and \( D_f \) denote the data modality and data features, respectively, as considered in the given condition. In the second phase, a nested iteration evaluates filtered candidate services from stage one $M_f$ against QoS parameters. The algorithm ensures that the selected services $M_f$ exhibit better QoS features to filter the services that are lower the $M_u^c$ (\textit{Algorithm \ref{alg2}, lines 10–13}). This step refines the selection by ensuring that only services that perform competitively better than the underperforming ones are considered for final selection. The algorithm's output is the set \( M' \), which contains the most suitable candidate MLaaS services.



\begin{algorithm}
\caption{Candidate MLaaS Selection Model}\label{alg2}
\begin{algorithmic}[1]
\State \textbf{Input:} \( M = \{M_1, M_2, ..., M_n\} \), \( M_u = \{M_p, M_q, M_r\} \)
\State \textbf{Output:} Candidate MLaaS Services \( M' \)
\State Initialize \( M' \gets \emptyset \) \Comment{Filtered MLaaS services}
\State Initialize \( M_u^c \) \Comment{Initiate \( M_u^c \) composition}
\For{$i \in M$}
    \If{$M_u^c\{\mathcal{N}_{\Phi},D_m, D_f\} \sim M_f\{\mathcal{N}_{\Phi},D_m, D_f\}$}  
        \State Append $M_i$ to $M_f$
    \EndIf
\EndFor
\For{$i \in M_f$}
    \For{$j \in M_u$}
        \If{$M_f\{E_f, Q, D_v\} > \text{mean}(M_u^c\{E_f, Q, D_v\})$}  
            \State Append $M_i$ to $M'$
        \EndIf
    \EndFor
\EndFor
\State \Return \( M' \)
\end{algorithmic}
\end{algorithm}

\subsection{Adaptive Composition Rules} 
We introduce adaptive MLaaS composition rules derived from the principles of Machine Learning (ML) and Collaborative Learning (CL) \cite{zhao2018federated, huang2022tackling, fu2023client, lai2021oort, zhang2023multi}. First, we define functionally dependent composition rules to assess data and model heterogeneity, considering attributes such as weights and data distributions across labels. We also establish QoS-dependent composition rules, which assess service alignment based on key quality factors, including reliability, efficiency, and cost. These rules compute a confidence score in the range (0,1) to determine the adaptive composition. We formulate five composition rules, defined as follows:

\textbf{a) Data Utility Measurement (\(DUM\))}:  
DUM evaluates the data distribution of an MLaaS service across data features, represented as \(D_f = \{1, 2, ..., c\}\). Data distribution is crucial in assessing ML compatibility, as variations can significantly affect ML performance. We define the Euclidean distance (\(ED\)) to quantify this similarity, a widely recognized metric for measuring distributional divergence and guiding client selection in CL \cite{zhao2018federated}. The \( ED(h_p^c, h^i) \) data distributions between a candidate MLaaS and an performing composition  can be computed as follows:
\vspace{2mm}
\begin{equation}\label{DUM}
DUM_{M_p^c, M'} =
\begin{cases}
1 & \text{if }  \sqrt{\sum_{i=1}^{n} (h_p^c - h_i')^2} < \theta_d,  \\
0 & \text{otherwise}.
\end{cases}
\vspace{-2mm}
\end{equation}

\( h = (h_1, \ldots, h_n) \) represents dataset distributions across \( n \) class labels, where each \( h_i \) is the sample count for a class. The \( ED(H_p^c, H') \) measures distribution skew, with lower values indicating better alignment with the existing composition. A tunable threshold \( \theta_d \) filters out services with high skew, ensuring model consistency.

\textbf{b) Model Utility Measurement (\(MUM\))}:  
The divergence between the weight updates of a candidate MLaaS (\( M_i \)) and the existing composition (\( M_p^c \)) can be determined using MUM. In CL settings \cite{fu2023client}, model weights (\( w \)) represent learned parameters, and significant divergence indicates a meaningful update to the system. The MUM between the existing composition (\( M_p^c \)) and the candidate MLaaS (\( M_i \)) is defined as: 

\begin{equation}\label{mum}
\text{MUM}_{M_p^c, M'} = \frac{1}{|w|} \sum_{i=1}^{|w|} \left| \frac{w'_{i} - \bar{w}_p^c}{\bar{w}_p^c} \right|.
\end{equation}

Here, \( \bar{w}_p^c \) and \( w \) denote the weight parameters of the existing composition and the candidate service, respectively. The term \( w'_i \) represents the \( i \)th weight in the existing composition. A low divergence value indicates minimal updates from the candidate MLaaS, making it insignificant.



\textbf{c) Scalability Measurement (SM):}  
Response time determines whether a candidate service \( M' \) can be integrated into an existing composition framework without exceeding acceptable limits. Each service is represented by a response-time vector \( \mathbf{t} = \{ t_1, t_2, \ldots, t_n \} \), where \( t_i \) denotes the response time for task \( i \). The composition must ensure efficient execution by preventing excessive latency. Studies such as \cite{lai2021oort} have explored strategies for measuring latency overhead. Following this, we define the SM as follows:

\begin{equation}\label{rtc}
SM_{M_p^c, M'} =
\begin{cases}
1 & \text{if } (T / t')^{\alpha} < \theta_T, \\
0 & \text{otherwise}.
\end{cases}
\end{equation}

Here, \( T \) and \( t' \) denote the response times of the existing composition and the candidate service, respectively. The penalty factor \( \alpha \) penalizes services with excessive latency. If the combined response time exceeds \( \theta_T \), the candidate service is deemed unsuitable due to its negative impact on system performance and efficiency.

\textbf{d) Historical Quality Score (\(HQS\)):}  
The HQS estimates the historical performance of a candidate service \( M_i \) relative to a performing composition \( M_p^c \). Each service maintains a historical quality vector \( q = (q_1, q_2, \ldots, q_n) \), where \( q_i \) denotes the past quality score for a specific metric. The historical quality score for task \( i \) is computed as:
\begin{equation}\label{hqu}
\text{HQU}_{M_p^c, M'} = q_{\text{task}_i}
\begin{cases}
1, & \text{if } \text{sim}(q_{\text{task}_p^c}, q_{\text{task}_i}) \geq \theta_q, \\
0, & \text{otherwise}.
\end{cases}
\end{equation}
Where \( TR \) is the set of trial rounds.  A higher similarity score (\( \text{HQU} > \theta_q \)) suggests that the candidate service shares historical performance patterns with the composition, ensuring smooth integration.

\textbf{e) Service Reliability Score (\(SRS\)):}  The SRS evaluates the consistency of a candidate service \( M_i \) in completing assigned tasks within an performing composition \( M_p^c \). Each service has a reliability score vector \( \mathbf{r} = \{ r_1, r_2, \ldots, r_n \} \), where \( r_i \) is computed for every completed task. The service reliability score is defined as follows:

\begin{equation}\label{sru}
\text{SRU}_{M_p^c, M'} = \frac{1}{|Task|} \sum_{t \in T} r_t
\begin{cases}
1 & \text{if } \text{SRS} \geq \theta_r, \\
0 & \text{otherwise}.
\end{cases}
\end{equation}

Here, \( \theta_r \) is the reliability threshold. A candidate service with \( \text{SRS} \geq \theta_r \) is considered stable and suitable for the existing composition plan, whereas lower values indicate potential failure risks.

\begin{algorithm}
\caption{Adaptive MLaaS Composition Model}\label{alg3}
\begin{algorithmic}[1]
\State \textbf{Input:} \( M', M_p^c = \{w, D_f,E_f, Q, t, R_t \},  \theta_S, \theta_R, \theta_T, \theta_Q, \lambda^s \)
\State \textbf{Output:} MLaaS Confidence Scores
\State $\mathbf{CS} \gets 0$ \Comment{Initialize confidence scores}
\For{all $M_i \in M'$}
    \State $DUM \gets 1$ \textbf{if} $\text{ED}(H_p^c, H_i') < \theta_d$ \textbf{else} $0$
    \State $MUM \gets 1$ \textbf{if} $\text{MUM}(W_p^c, W_i') > \theta_S$ \textbf{else} $0$
     \State $SM \gets 1$ \textbf{if} $\left(\frac{T}{t'}\right)^\alpha < \theta_T$ \textbf{else} $0$
    \State $HQU \gets 1$ \textbf{if} $\text{sim}(q_{\text{task}_p^c}, q_{\text{task}_i}) \geq \theta_q$ \textbf{else} $0$
     \State $SRU \gets 1$ \textbf{if} $\frac{1}{|Task|} \sum_{t \in T} r_t \geq \theta_r$ \textbf{else} $0$
    \State $S_i\gets [DUM, MUM, SRU, SM, HQU]$
    \State $\text{CS}_i \gets \sum_{j=1}^{5} \lambda^s \cdot s_j$ \Comment{Weighted confidence score}
    \State Append $\text{CS}_i$ to $\mathbf{CS}$
\EndFor
\State \Return $\mathbf{CS}$
\end{algorithmic}
\end{algorithm}
\vspace{-2mm}
\subsection{Adaptive MLaaS Composition Model}
We propose a novel adaptive MLaaS composition model to determine the composition between \( M_p^c\), and candidate service \( M'\). Algorithm~\ref{alg3} demonstrates adaptive MLaaS composition designed to evaluate the composition. The algorithm evaluates compatibility across multiple dimensions, including data heterogeneity, model usability, system reliability, efficiency and quality factors. The algorithm takes as input a set of candidate MLaaS services \( M'\), performing composition \( M_p^c\), along with predefined threshold parameters \( \theta_S, \theta_R, \theta_T, \theta_Q\), and the weight factor \( \lambda^s \)(\textit{Algorithm \ref{alg3}, lines 1–2}). These factors collectively determine the \textit{Confidence Scores (CS)}, which serve as the output (\textit{Algorithm \ref{alg3}, lines 3}). First, the model evaluates adaptive compatibility rules to determine composition, focusing on data and model heterogeneity, as well as system specifications attributes, including efficiency, latency, and quality factors (\textit{Algorithm \ref{alg3}, lines 4–10}). The performance weight \( \lambda^s \) is used to calculate the confidence scores based on individual composition rules. It allows IoT providers to customize and define their preferences. The computed \textbf{CS} values represent the confidence of integrating a candidate service into the adaptive composition framework without recomposition.(\textit{Algorithm \ref{alg3}, lines 11–14})  

\subsection{Contextual MAB-based Adaptive MLaaS Composition}

Adaptive MLaaS composition presents a combinatorial optimization challenge, as more than one service may not function effectively within a given composition. Due to the dynamic nature of IoT,  MLaaS services exhibit varying functional and non-functional attributes. To address this, we structured the optimization problem as a contextual multi-armed bandit (CMAB) problem. In this framework, the player (i.e., the IoT provider) evaluates the contextual features of all candidate MLaaS services (arms) and selects the most suitable arm. An arm represents an individual MLaaS service, characterized by its functional and non-functional attributes. Each selected arm receives a corresponding reward. 
The objective of the problem is to select an arm from multiple arms with their contextual features so that the total cumulative reward is maximized. Contextual features, reward function and UCB are the key components defines as following



\subsubsection{Contextual Feature}  
Contextual features are crucial because arm rewards depend on both adaptive composition and external functional and QoS factors. These factors enable the contextual MAB to adapt dynamically, improving decision quality. The contextual features of candidate \( i \in M' \) are given by:  

\begin{equation}
    C_{f,i} = [ D_I, Qos_i ], \quad \forall i \in M'
\end{equation}
where \( D_i \) represents data specifications, and \( Qos_i \) includes efficiency \( A_i \), latency \( P_i \), reliability \( R_i \), and quality factor \( Q_i \).  

\subsubsection{Reward function}
The adaptive MLaaS composition model defines the reward function to evaluate the composition of the candidate services, denoted as follows:  
\begin{equation}
    r_t{i} = \sum_{i \in \mathcal{N}} \cdot \text{CS}_{i}, \quad \forall i \in \mathcal{N}, \mathbb{E}[r_t{i} | C_{f,i}] = C_{f,i}^{T} \theta^*
\end{equation}


However, the correlation between a candidate service’s \textbf{confidence score} and its \textbf{contextual features} remains unknown. By continuously refining this relationship, the algorithm enhances candidate selection, ensuring efficient and stable adaptive composition. Following similar approaches \cite{li2010contextual, huang2022contextfl}, we model the expected reward function as a function of contextual features \( C_{f,i} \). To estimate this correlation, we define the coefficient vector as \( \hat{\theta}_{i} = A^{-1}_{i} b_{i} \), where \( A_i \) is a decision matrix initialized as an identity matrix, and \( b_i \) starts as a zero matrix. These matrices update iteratively with new observations, refining the estimated correlation between contextual features and confidence scores. This enables the CMAB framework to dynamically adapt and select the most suitable MLaaS services with high confidence. The CMAB-based Adaptive MLaaS Composition Algorithm (\textit{Algorithm~\ref{alg4}})optimizes MLaaS selection by dynamically choosing services that maximize composition while balancing exploration and exploitation. The algorithm starts with a set of candidate MLaaS services \( M' \), an exploration-exploitation parameter \( \alpha \), a performing composition \( M_c^p \), and the number of top \( K \) services to be selected. The decision matrices are initialized as \( A_i \gets I_d \) and \( b_i \gets 0_d \) to provide a baseline for learning (\textit{Algorithm~\ref{alg4}, lines 1–4}). At each iteration, the algorithm observes the contextual features \( C_{f,i} \) of all candidates. Min-Max Scaling is applied to normalize these features to \([0,1]\), ensuring consistency in evaluation (\textit{Algorithm~\ref{alg4}, lines 6–9}).  The Upper Confidence Bound (UCB) score \cite{li2010contextual} is computed for each candidate. The coefficient vector is estimated as \( \hat{\theta}_i \gets A_i^{-1} b_i \), refining selection based on past observations (\textit{Algorithm~\ref{alg4}, line 11}). The computed UCB scores defined ad follows:


\begin{equation}
p_{t,G} = \sum_{i=1}^{K} \sigma \left( C_{f,i}^{T} \hat{\theta}_i + \alpha \sqrt{C_{f,i}^{T} A_i^{-1} C_{f,i}} \right)
\vspace{-2mm}
\end{equation}

(\textit{Algorithm \ref{alg4}, lines 12–14}), ensuring that the top \( K \) optimized service combinations are selected based on their UCB scores. The highest-scoring combination is then added to the set of chosen arms and undergoes an adaptive MLaaS composition evaluation. The resulting confidence scores serve as rewards in the next iteration (\textit{Algorithm \ref{alg4}, lines 14–18}). To refine selection accuracy, the decision matrices \( A_i \) and \( b_i \) are updated (\textit{Algorithm \ref{alg4}, lines 21–24}), enabling the algorithm to learn and adapt dynamically. Over multiple iterations, this approach efficiently explores and exploits MLaaS service combinations, ultimately selecting the most optimal services based on accumulated confidence scores.


{\scriptsize
\begin{algorithm}
\caption{CMAB-based Adaptive MLaaS Composition}\label{alg4}
\begin{algorithmic}[1]
\State \textbf{Input:} $M = \{M_1, M_2, ..., M_i\}$,$\alpha$,$M_c'$, $K$
\State \textbf{Output:} Optimal MLaaS selection set $Arms$
\State Initialize $A_i \gets I_d$, $b_i \gets 0_d$, $\forall i \in \mathcal{N}$
\For{each selection round $t = 1$ to $T$}
    \State Observe contextual features $C_{f,i}$ $i \in \mathcal{N}$
    \State $S \gets \mathcal{N}$, $Arms \gets \{\}$
    \For{$k = 1$ to $K$}
        \State Randomly select $\lfloor \frac{|\mathcal{N}|}{K} \rfloor$ candidates from $S$
        \State $C_{f,i} \gets \text{MinMaxScale}(C_{f,i})$
        \State Compute coefficient vector: \quad $\hat{\theta}_i \gets A_i^{-1} b_i$
        \State UCB \quad $ p_{t,G} = \sum_{i=1}^{K} \left( C_{f,i}^{T} \hat{\theta}_i + \alpha \sqrt{C_{f,i}^{T} A_i^{-1} C_{f,i}} \right)$
        \State Select candidate: $a_i \gets \arg\max p_{t,G}$
        \State Remove $a_i$ from $S$ and add it to $Arms$
    \EndFor
    \State Execute action $Arms$ to select $K$ MLaaS services
    \For{$i = 1$ to $Arms$}
     \State Adaptive MCM CS(i,$M_c'$)
     \State \Return average reward $r_ti$
    \EndFor
    \For{each $i \in Arms$}
        \State Update decision matrices:
        \State \quad $A_i \gets A_i + C_{f,i} C_{f,i}^{T}$
        \State \quad $b_i \gets b_i + R_{t,i} C_{f,i}$
    \EndFor
\EndFor
\State \Return $Arms$
\end{algorithmic}
\end{algorithm}}
\vspace{-3mm}
\section{Experiment Results and Discussion}\label{Experiment}


A series of experiments are designed to evaluate the proposed framework. First, we analyze the efficiency of the proposed adaptive MLaaS composition. Since no existing models offer a direct comparison, we create a benchmark by recomposing the MLaaS. Additionally, we compare its performance with traditional QoS-based \cite{ardagna2007adaptive} and rule-based adaptive \cite{wang2010adaptive} composition techniques. To assess scalability, we evaluate the CMAB-based adaptive MLaaS composition against state-of-the-art methods, including brute-force \cite{lakhdari2020composing}, epsilon-greedy \cite{liu2021improving}, and genetic algorithms \cite{tan2014automated}. Finally, we analyze the time complexity of the service assessment model relative to LOO \cite{wang2019measure}, NCS \cite{zhang2024towards}, and SHAP \cite{song2019profit}. All experiments were executed on an Intel Core i7 (16GB RAM) using Python, with results and source code available in our repository\footnote{https://anonymous.4open.science/r/AdaptiveMLaaS-2612}.


\subsection{Experiment Setup}
To the best of our knowledge, there is no existing dataset about MLaaS Services in the IoT environment. In addition, MLaaS providers often keep incomplete information on their services, offering only limited details through publicly available sources. To address this gap, we generated MLaaS data by setting up an MLaaS composition using a publicly available MNIST \cite{cohen2017emnist}, FEMINIST\cite{xiao2017fashion} AND HAR datasets\cite{misc_pamap2_physical_activity_monitoring_231}. we further conducted data mining to extract insights from leading MLaaS advertisement platforms. This process involved collecting data from ten major providers, such as Amazon SageMaker, Google AI Platform, Microsoft Azure Machine Learning, IBM Watson Studio, and Oracle Cloud Infrastructure Data Science. The dataset is distributed across 20 clients and configured for both IID and non-IID setups to simulate real-world MLaaS datasets exhibiting variations in data distributions, system configurations, and model architectures. To simulate real-world dynamics, we introduce concept drift and data drift scenarios, creating a dynamic IoT environment. The dataset consists of both functional and non-functional attributes, including model parameters (weights, gradients), data specifications (modality, features, volume), and QoS metrics such as accuracy (65--99\%) and latency (5--100 ms per sample).


\begin{figure}[ht]
    \centering
    \includegraphics[width=1\columnwidth]{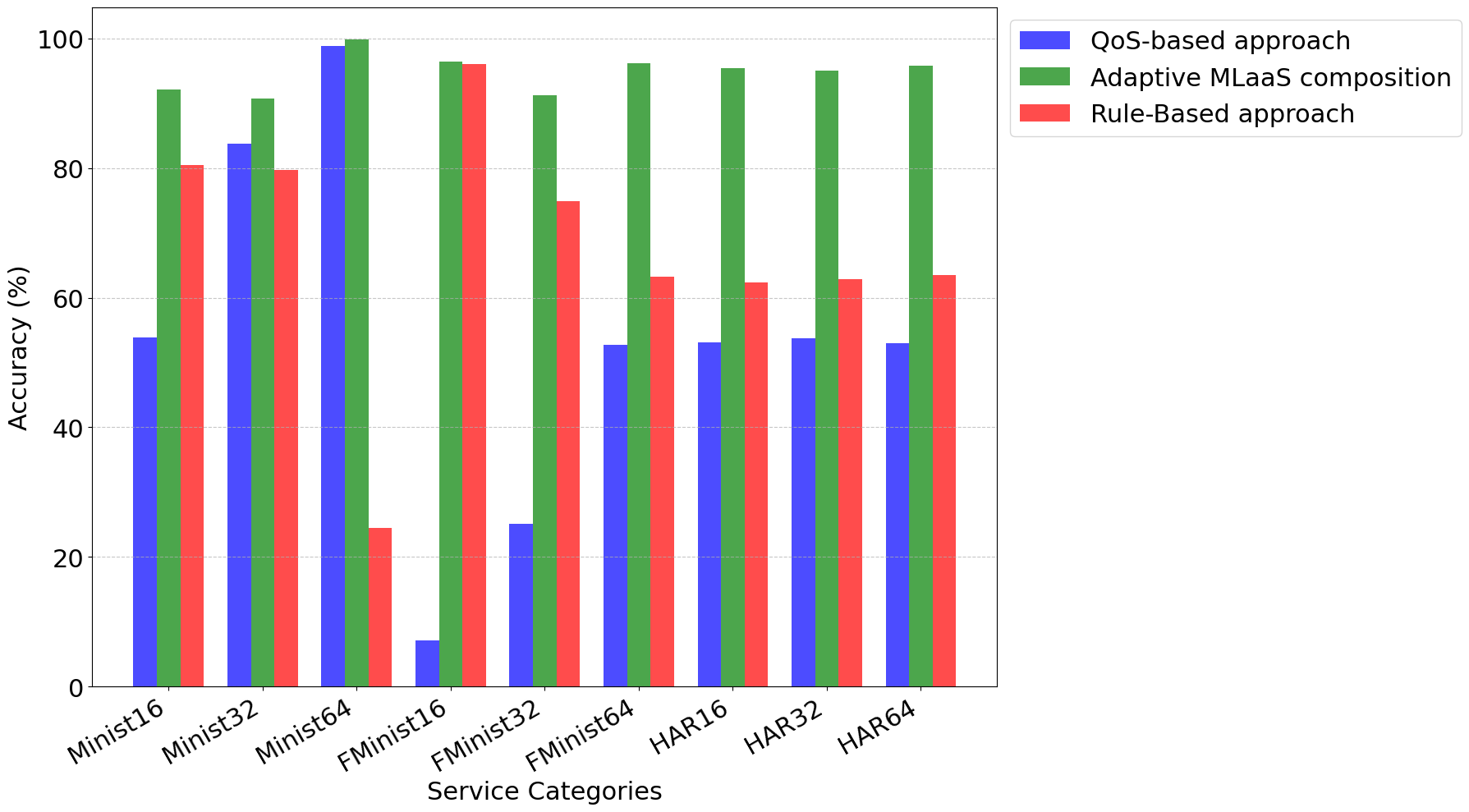}
    \caption{Efficiency of Adaptive MLaaS Composition}
    \label{fig:exp2}
    \vspace{-5mm}
\end{figure}

\subsection{Experiment 1: Efficiency of the Adaptive MLaaS Composition Model}
We investigate the efficiency of the proposed adaptive MLaaS composition model by performing recomposition, which serves as the baseline approach. Existing studies lack models to perform a direct comparison of adaptive MLaaS composition. However, recomposition is computationally expensive and time-consuming. To address this challenge, the proposed adaptive composition model efficiently determines the composition of candidate services within the existing composition without recomposing. Moreover, we compare its efficiency with traditional QoS-based \cite{ardagna2007adaptive} and rule-based adaptive composition techniques.
To evaluate adaptive composition, we perform experiments using three different datasets, MNIST, FMNIST, and HAR, across varying model architectures. 
\textit{Figure~\ref{fig:exp2}} presents the accuracy comparison, showing that the adaptive composition model consistently outperforms the QoS-based and rule-based approaches across all model categories. The adaptive composition model consistently achieves the highest accuracy of 97\%, followed by the rule-based approach with 78\%, while the QoS-based method lags with an average accuracy of 55\%. The results highlight the influence of the dataset specifications' characteristics and the model's size on the composition's accuracy, reinforcing the importance of adaptive selection. Table~\ref{Evaluation} demonstrates the effectiveness of adaptive MLaaS across different model sizes and datasets. The adaptive approach consistently outperforms others, achieving over 95\% accuracy across datasets, confirming its robustness in determining optimal composition. QoS-based composition and rule-based approach struggles with small models, notably on FMNIST16 (26.21\% accuracy) and  MNIST16 (44.47\% accuracy). Notably, larger models generally yield better performance across all methods, reinforcing the impact of model capacity on adaptive composition effectiveness.

\begin{table}[ht]
\centering
\caption{Performance Metrics Across Different Datasets}
\label{Evaluation}
\renewcommand{\arraystretch}{1.5} 
\setlength{\tabcolsep}{6pt} 
\large 
\resizebox{\columnwidth}{!}{%
\begin{tabular}{l|ccc|ccc|ccc}
\hline
\multirow{2}{*}{\textbf{Method}} 
& \multicolumn{3}{c|}{\textbf{Accuracy (\%)}} 
& \multicolumn{3}{c|}{\textbf{Precision (\%)}} 
& \multicolumn{3}{c}{\textbf{Recall (\%)}} \\ 
\cline{2-10}
& \textbf{MNIST16} & \textbf{FMNIST16} & \textbf{HAR16}  
& \textbf{MNIST32} & \textbf{FMNIST32} & \textbf{HAR32}  
& \textbf{MNIST64} & \textbf{FMNIST64} & \textbf{HAR64}  \\ \hline
QoS-based Composition \cite{ardagna2007adaptive}   
& \Large 88.82  & \Large 26.21  & \Large53.28  
& \Large90.77  & \Large56.29  & \Large69.87  
& \Large73.96  & \Large51.31  & \Large55.93  \\ 
\textbf{Proposed Adaptive MLaaS}  
& \textbf{\Large 96.75}  & \textbf{\Large 94.20}  & \textbf{\Large 95.40}  
& \textbf{\Large 96.29}  & \textbf{\Large 92.33}  & \textbf{\Large 95.36}  
& \textbf{\Large 93.59}  & \textbf{\Large 91.22}  & \textbf{\Large 95.41}  \\ 
Rule-Based Composition  \cite{wang2010adaptive}  
& \Large44.47  & \Large78.95  & \Large62.90  
&\Large 56.86  & \Large70.09  & \Large66.10  
& \Large58.99  & \Large63.19  & \Large61.21  \\ \hline
\end{tabular}}
\vspace{-4mm} 
\end{table}

\subsection{Experiment 2: Scalability of CMAB-based adaptive MLaaS composition}
\begin{figure}[ht]
    \centering
    \includegraphics[width=0.9\columnwidth]{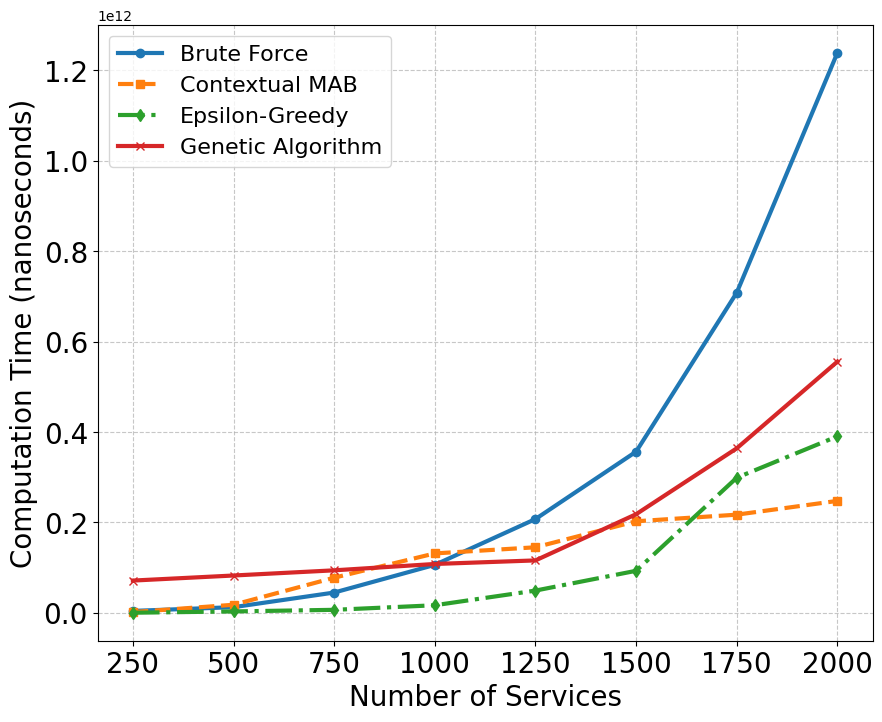}
    \caption{Scalability of CMAB-based Adaptive MLaaS Composition}
    \label{fig:exp31}
    \vspace{-5mm}
\end{figure}

We investigate the \textit{scalability} of the proposed \textit{CMAB based adaptive composition algorithm} against the other state-of-the-art model brute force \cite{lakhdari2020composing}, epsilon-greedy \cite{liu2021improving} and genetic algorithms \cite{tan2014automated}.  \textit{Figure \ref{fig:exp31}} illustrates the execution time of the \textit{CMAB based adaptive composition} against the growing number of MLaaS services. As the number of MLaaS services grows, the number of possible combinations required to compose these services increases significantly over time. The results demonstrate that the CMAB algorithm significantly outperforms the brute-force approach regarding computational efficiency, especially as the service size scales. The brute-force method shows an exponential increase in computation time, reaching over 1.2 ns due to exploring all possible combinations. The results highlight the contextual MAB's effectiveness in navigating the search space efficiently, identifying viable service compositions without the need for exhaustive exploration. Compared to other methods, such as Epsilon-Greedy and the Genetic Algorithm, the contextual MAB consistently demonstrates faster execution times, particularly as the number of services exceeds 2000. These results underscore the scalability and practical applicability of the contextual MAB algorithm for efficiently managing large-scale MLaaS composability tasks.

\subsection{Experiment 3: Service Assessment Model Time Efficiency}

\begin{figure}[ht]
    \centering
    \includegraphics[width=0.9\columnwidth]{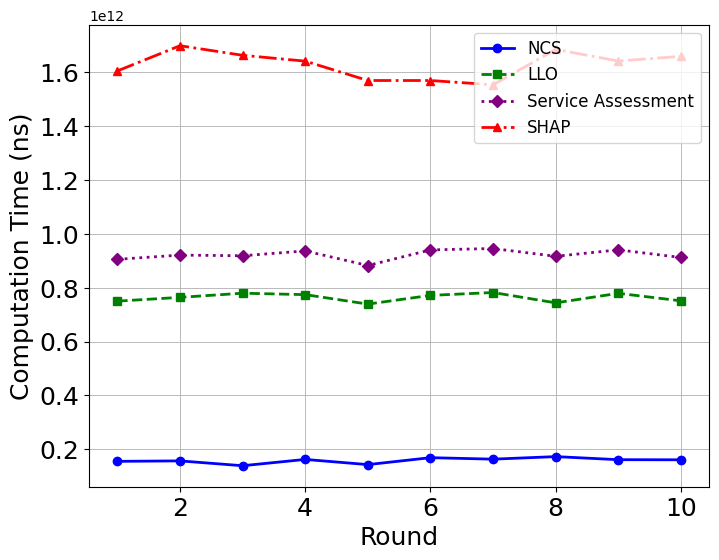}
    \caption{Computation Time Comparison of SAM}
    \label{exp1}
    \vspace{-5mm}
\end{figure}

We evaluate the time complexity of the \textit{service assessment model} against three traditional techniques: LOO, NCS, and SHAP. \textit{Figure~\ref{exp1}} presents the execution time in ns across multiple rounds. Theoretical analysis shows that \textit{SHAP} has a complexity of \( O(n^2) \), whereas \textit{LOO} and \textit{NCS} operate with a lower complexity of \( O(n) \).  The proposed \textit{Service Assessment Score}, which integrates both functional and QoS attributes while balancing LOO and NCS, also follows \( O(n) \) complexity. Among these methods, SHAP incurs the highest computational cost, averaging \textit{1.7 ns per round} due to its exhaustive evaluation of all possible client combinations. In contrast, \textit{NCS} and \textit{LOO} achieve significantly lower average computation times of \textit{0.3 and 0.7 ns per round}, respectively. However, both \textit{LOO} and \textit{NCS} do not effectively incorporate functional and QoS attributes. To address this, we propose the \textit{Service Assessment Model}, which demonstrates a strong impact in identifying underperforming services while maintaining an average execution time of \textit{0.9 ns per round}. The graph shows the effectiveness of the proposed method in reducing computational overhead while maintaining reliable assessments.

\subsection{Discussion}

Our experiments show that adaptive composition effectively lowers both the cost and complexity of recomposition. This ensures that compositions remain stable and efficient over time rather than being discarded and recomposed from scratch. However, one key limitation of our framework is that while it effectively manages optimal services, its efficiency in identifying trade-off services is slightly lower. Addressing this issue requires a deeper investigation into the underlying features of the possible trade-off services. One approach could be to collect more samples of such cases and develop a set of correlation-based heuristics to refine the decision-making process. Moreover, the framework currently relies on a one-to-one replacement strategy, where each underperforming service is substituted with an alternative. Due to this approach, the framework requires more computations and must run extended optimization cycles for adaptation. Investigating whether multiple underperforming services can be replaced with a single, more efficient service could potentially reduce computational costs and improve optimization efficiency. Exploring such alternatives could lead to a more streamlined and adaptive composition process.

\section{Conclusion}
In this paper, we propose an adaptive MLaaS composition framework for IoT environments. The proposed framework consists of three key components (1) identification of underperforming services, (2) identification of candidate services, and (3) evaluation and optimization of the adaptive composition. First, we design a service assessment model that measures the contribution of each service to identify underperforming services. Next, we develop an MLaaS selection model that utilizes both underperforming and performing services to filter the potential candidates. We propose an adaptive MLaaS composition model to determine the best replacement instead of fully recomposing. The experimental results demonstrate that the proposed adaptive MLaaS composition achieves 95\% accuracy compared to the traditional rule-based and QoS-based composition. We formulate the adaptive composition problem as a contextual optimization problem and introduce a CMAB algorithm to update the composition. Results show that CMAB is more scalable than brute-force search, epsilon-greedy, and genetic algorithms. In this study, we assume that replacing an underperforming service helps rebuild adaptive compositions to meet IoT requirements. However, future work will explore integrating new services into existing compositions rather than relying solely on direct replacements. Additionally, we will investigate whether removing an underperforming service without substitution can still satisfy IoT provider requirements. These aspects will further refine the composition process, enhancing adaptability while ensuring efficiency in dynamic IoT environments.

\bibliographystyle{ieeetr}
\bibliography{Bibliography1}

\begin{thebibliography}{10}

\bibitem{lee2015internet}
I.~Lee and et~al., ``The internet of things (iot): Applications, investments, and challenges for enterprises,'' {\em Business Horizons}, pp.~431--440, 2015.

\bibitem{bianchi2019iot}
V.~Bianchi and et~al., ``Iot wearable sensor and deep learning: An integrated approach for personalized human activity recognition in a smart home environment,'' {\em IEEE IoTJ}, pp.~8553--8562, 2019.

\bibitem{ribeiro2015mlaas}
M.~Ribeiro and et~al., ``Mlaas: Machine learning as a service,'' in {\em 2015 IEEE ICMLA}, pp.~896--902, IEEE, 2015.

\bibitem{noshiri2021machine}
N.~Noshiri and et~al., ``Machine learning-as-a-service performance evaluation on multi-class datasets,'' in {\em 2021 IEEE SmartIoT}, pp.~332--336, IEEE, 2021.

\bibitem{xie2024skyml}
S.~Xie and et~al., ``Skyml: A mlaas federation design for multicloud-based multimedia analytics,'' {\em IEEE TMM}, 2024.

\bibitem{fadlullah2018delay}
Z.~M. Fadlullah and et~al., ``On delay-sensitive healthcare data analytics at the network edge based on deep learning,'' in {\em 2018 14th IWCMC}, pp.~388--393, IEEE, 2018.

\bibitem{muccini2020leveraging}
H.~Muccini and et~al., ``Leveraging machine learning techniques for architecting self-adaptive iot systems,'' in {\em 2020 IEEE SMARTCOMP}, pp.~65--72, IEEE, 2020.

\bibitem{ma2011adaptive}
Z.~Ma and et~al., ``Adaptive service composition based on runtime requirements monitoring,'' in {\em 2011 IEEE ICWS}, pp.~339--346, IEEE.

\bibitem{wang2010adaptive}
H.~Wang and et~al., ``Adaptive and dynamic service composition using q-learning,'' in {\em 2010 22nd IEEE ICTAI}, 2010.

\bibitem{ardagna2007adaptive}
D.~Ardagna and et~al., ``Adaptive service composition in flexible processes,'' {\em IEEE TSE}, pp.~369--384, 2007.

\bibitem{song2019profit}
T.~Song and et~al., ``Profit allocation for federated learning,'' in {\em 2019 IEEE Big Data}, pp.~2577--2586, IEEE, 2019.

\bibitem{wang2019measure}
G.~Wang and et~al., ``Measure contribution of participants in federated learning,'' in {\em 2019 IEEE BigData}, pp.~2597--2604, IEEE, 2019.

\bibitem{zhao2018federated}
Y.~Zhao and et~al., ``Federated learning with non-iid data,'' {\em arXiv preprint arXiv:1806.00582}, 2018.

\bibitem{lai2021oort}
F.~Lai and et~al., ``Oort: Efficient federated learning via guided participant selection,'' in {\em 15th $\{$USENIX$\}$ Symposium on Operating Systems Design and Implementation ($\{$OSDI$\}$ 21)}, pp.~19--35, 2021.

\bibitem{tay2019xylorix}
Y.~Tay, ``Xylorix: An ai-as-a-service platform for wood identification,'' in {\em IAWA-IUFRO}, pp.~20--22, 2019.

\bibitem{pohl2018data}
M.~Pohl and et~al., ``A data-science-as-a-service model.,'' in {\em CLOSER}, pp.~432--439, 2018.

\bibitem{zhang2024towards}
M.~Zhang and et~al., ``Towards fair, robust and efficient client contribution evaluation in federated learning,'' {\em arXiv preprint arXiv:2402.04409}, 2024.

\bibitem{huang2022tackling}
J.~Huang and et~al., ``Tackling mavericks in federated learning via adaptive client selection strategy,'' in {\em AAAI}, vol.~2023, 2022.

\bibitem{fu2023client}
L.~Fu and et~al., ``Client selection in federated learning: Principles, challenges, and opportunities,'' {\em IEEE IoTJ}, 2023.

\bibitem{zhang2023multi}
M.~Zhang and et~al., ``Multi-criteria client selection and scheduling with fairness guarantee for federated learning service,'' {\em arXiv preprint arXiv:2312.14941}, 2023.

\bibitem{wang2014adaptive}
H.~Wang and et~al., ``Adaptive and dynamic service composition via multi-agent reinforcement learning,'' in {\em 2014 IEEE ICWS}, pp.~447--454, IEEE, 2014.

\bibitem{tan2014automated}
T.~H. Tan, M.~Chen, {\'E}.~Andr{\'e}, J.~Sun, Y.~Liu, and J.~S. Dong, ``Automated runtime recovery for qos-based service composition,'' in {\em 23rd international conference on WWW}, pp.~563--574, 2014.

\bibitem{ebron2023fedtruth}
S.~C. Ebron~Jr and K.~Yang, ``Fedtruth: Byzantine-robust and backdoor-resilient federated learning framework,'' {\em arXiv preprint arXiv:2311.10248}, 2023.

\bibitem{li2010contextual}
L.~Li and et~al., ``A contextual-bandit approach to personalized news article recommendation,'' in {\em IWWW}, pp.~661--670, 2010.

\bibitem{huang2022contextfl}
H.~Huang, R.~Li, J.~Liu, S.~Zhou, K.~Lin, and Z.~Zheng, ``Contextfl: Context-aware federated learning by estimating the training and reporting phases of mobile clients,'' in {\em 2022 IEEE ICDCS}, pp.~570--580.

\bibitem{lakhdari2020composing}
A.~Lakhdari and et~al., ``Composing energy services in a crowdsourced iot environment,'' {\em IEEE TCS}, vol.~15, no.~3, pp.~1280--1294, 2020.

\bibitem{liu2021improving}
Y.~Liu and et~al., ``Improving ant colony optimization algorithm with epsilon greedy and levy flight,'' {\em CIS}, vol.~7, no.~4, pp.~1711--1722, 2021.

\bibitem{cohen2017emnist}
G.~Cohen and et~al., ``Emnist: Extending mnist to handwritten letters,'' in {\em 2017 international joint conference on neural networks (IJCNN)}, pp.~2921--2926, IEEE, 2017.

\bibitem{xiao2017fashion}
H.~Xiao and et~al, ``Fashion-mnist: a novel image dataset for benchmarking machine learning algorithms,'' {\em arXiv preprint arXiv:1708.07747}, 2017.

\bibitem{misc_pamap2_physical_activity_monitoring_231}
A.~Reiss, ``{PAMAP2 Physical Activity Monitoring}.'' UCI Machine Learning Repository, 2012.
\newblock {DOI}: https://doi.org/10.24432/C5NW2H.

\end{thebibliography}

\end{document}